\documentclass[twoside]{article}

\usepackage[accepted]{aistats2025}
\usepackage[table,xcdraw, dvipsnames]{xcolor}
\usepackage{graphicx}
\usepackage{svg}
\svgpath{{figures/}}
\usepackage{tikz}
\usetikzlibrary{
	arrows, arrows.meta, positioning, fit
}
\usepackage{caption}
\usepackage{subcaption}  % grids of images
\usepackage{mathtools,amsthm,amssymb}
\usepackage{bm}  % for bold greek math
\usepackage{mathrsfs}  % \mathscr command
\usepackage{algorithm}
\usepackage[noend]{algpseudocode}
\usepackage{thmtools}  % theorem, definition, and question lists

\theoremstyle{plain}
\newtheorem{theorem}{Theorem}[section]
\newtheorem{proposition}[theorem]{Proposition}

\theoremstyle{definition}

\theoremstyle{remark}

\usepackage[utf8]{inputenc}
\usepackage{enumerate}  % Custom numbered items, such as steps 1, 2, etc.
\usepackage{enumitem}  % Ordinals
\usepackage{moreenum}
\usepackage{verbatim}
\usepackage{hyperref}
\hypersetup{
    colorlinks=true,
    linkcolor=blue,
    filecolor=magenta,
    urlcolor=cyan,
    citecolor=blue,
    }
\usepackage{url}

\newcommand{\rev}[1]{#1}

\usepackage[round]{natbib}

\begin{document}

\twocolumn[

\aistatstitle{\textsc{SubSearch}: Robust Estimation and Outlier Detection for Stochastic Block Models via Subgraph Search}

\aistatsauthor{Leonardo Martins Bianco\textsuperscript{1} \And Christine Keribin\textsuperscript{1} \And Zacharie Naulet\textsuperscript{2}}

\aistatsaddress{}
%\aistatsaddress{Université Paris-Saclay, Inria \And Université Paris-Saclay, Inria  \And INRAE, Université Paris-Saclay}

]

\footnotetext[1]{Université Paris-Saclay, CNRS, Inria, Laboratoire de Mathématiques d’Orsay,
91405 Orsay, France}
\footnotetext[2]{Université Paris-Saclay, INRAE, MaIAGE, 78350, Jouy-en-Josas, France}

\begin{abstract}
    Community detection is a fundamental task in graph analysis, with methods often relying on fitting models like the Stochastic Block Model (SBM) to observed networks. While many algorithms can accurately estimate SBM parameters when the input graph is a perfect sample from the model, real-world graphs rarely conform to such idealized assumptions. Therefore, robust algorithms are crucial---ones that can recover model parameters even when the data deviates from the assumed distribution. In this work, we propose \textsc{SubSearch}, an algorithm for robustly estimating SBM parameters by exploring the space of subgraphs in search of one that closely aligns with the model's assumptions. Our approach also functions as an outlier detection method, properly identifying nodes responsible for the graph's deviation from the model and going beyond simple techniques like pruning high-degree nodes. Extensive experiments on both synthetic and real-world datasets demonstrate the effectiveness of our method.
\end{abstract}

\section{INTRODUCTION} \label{introduction}
% Old version of paragraph 1.
% Community detection on a graph is the task of partitioning its set of nodes in such a way that each partition, also called a community, can be interpreted as a group of nodes sharing some common property. This task arises naturally in a variety of fields, such as physics~\cite{PhysRevE.74.016110}, computer science~\cite{moore2017computersciencephysicscommunity}, and in the study of complex networks~\cite{bedi2016community}. One approach to community detection consists on fitting a generative model to the data observed. A popular generative model for graphs with communities is the Stochastic Block Model (SBM), introduced in~\cite{HOLLAND1983109} to study social networks. An interest in determining the information-theoretic and computational thresholds of this model led to a wide range of results and accompanying algorithms for community detection in the past decade, see~\cite{abbe2018community} for a comprehensive review.

% Bianca: Trocar as aplicações genéricas por aplicações mais interessantes.
Community detection on a graph is the task of partitioning its set of nodes in such a way that each partition, also called a community, represents a group of nodes sharing a common property, such as a similar pattern of connections. This problem has a wide range of applications, including understanding of the spread of epidemics~\citep{stegehuis2016epidemic}, customer segmentation in advertising~\citep{lalwani2015community}, and identification of criminals through online activity~\citep{sangkaran2020criminal}. One approach to community detection consists of fitting a model to the observed data, with the Stochastic Block Model (SBM)~\citep{HOLLAND1983109} being a popular choice. In the SBM, each node \(i\) is assigned a latent variable \(Z_i\) such that the sets \(\Omega_k = \{i : Z_i = k\}\), \(1 \leq k \leq K\), partition the graph into \(K\) communities. Any two nodes \(i\) and \(j\) are connected with a probability \(\Gamma_{Z_i Z_j}\), which depends only on their respective communities. Given a graph, the task of inferring the hidden partition \(\{\Omega_k\}_{k=1,\dotsc,K}\) is called community recovery (or retrieval), while estimating the connectivity parameters \(\Gamma\) is called parameter estimation. Common methods for these tasks include semidefinite programming (SDPs)~\citep{li2021convex}, spectral~\citep{lei_rinaldo}, and variational approaches~\citep{tabouy2020variational}.

% Old version of paragraph 2
% Despite this progress, many algorithms with guarantees for the SBM are highly sensitive to even small deviations from the model~\cite{cai2015robust}, suggesting that they rely on properties that are specific to the SBM and break down when the model is not perfectly accurate. This fragility poses a challenge for the practical use of these algorithms and opens an opportunity for the study of robust algorithms for community detection.

% Como melhorar:
% O que é model mispecification? --> DONE
% O que é adversarial perturbation? --> DONE
% O que exatamente são small deviations? --> REMOVIDO
% O que robustness significa? --> DONE

% Old reference to Abbe
% An interest in determining the information-theoretic and computational thresholds of the SBM led to a wide range of results and accompanying algorithms for community detection in the past decade, see~\cite{abbe2018community} for a comprehensive review. Despite this progress,

\begin{figure}
  \centering
  \includegraphics[width=\linewidth]{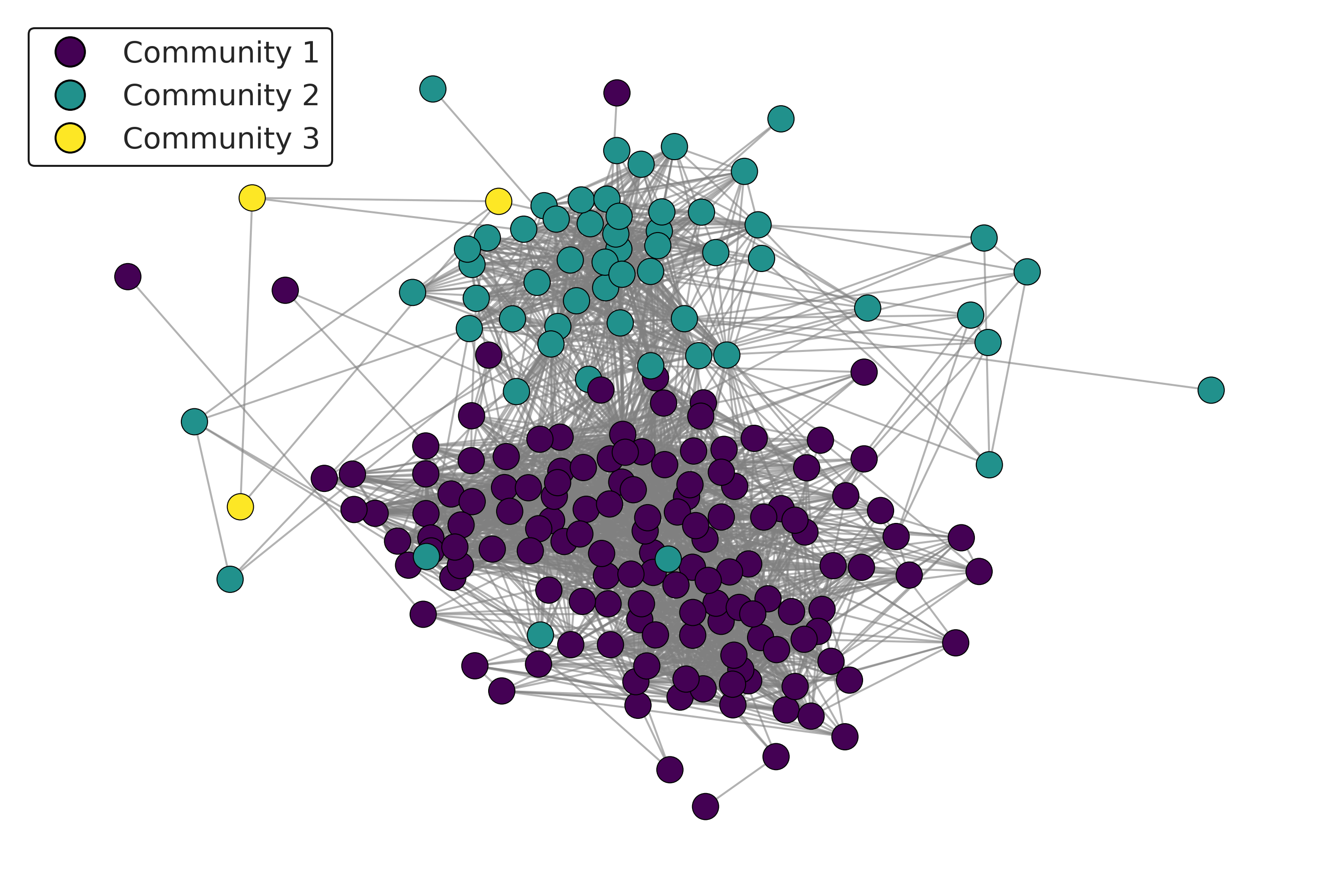}
  \caption{Spectral clustering applied to the jazz collaboration dataset~\citep{gleiser2003community}. Nodes represent jazz musicians, with edges being collaborations during 1912 - 1940. This algorithm, which has guarantees under the SBM~\citep{lei_rinaldo}, fails to separate the graph into its three main collaboration groups.}
  \label{fig:example_jazz}
\end{figure}

However, many algorithms that have guarantees under the assumption of an SBM are highly sensitive to model misspecification---when the input data is drawn from a distribution that differs from the model's assumptions~\citep{cai2015robust}. This sensitivity poses a significant challenge in practice, as real-world graphs rarely exhibit the level of connection homogeneity imposed by the SBM. As a result, these algorithms can produce inaccurate partitions or biased parameter estimates (see Figure \ref{fig:example_jazz}). One alternative is to use more complex models to describe the data (for example, a Degree-Corrected SBM~\citep{karrer2011stochastic}). While this works in some cases, it also introduces additional challenges, both theoretically and practically, due to the increased complexity of the model. Another solution, the one that interests us here, is to keep the simpler model and focus on developing robust algorithms---those producing reliable results even when given (slightly) misspecified inputs.

% The problem and its importance.
We consider the problem of robustly estimating connectivity parameters \(\Gamma\) under adversarial perturbations, a particular type of model misspecification in which an adversary modifies a sample from the (well-specified) model before passing it as input to the algorithm, with the goal of inducing as much error as possible. This problem is important for several reasons. First, knowing the community structure is insufficient for parameter estimation, as adversarial perturbations can significantly alter the observed connections within and between communities, distorting the true underlying probabilities. Second, identifying the outliers causing these perturbations is nontrivial. Third, several algorithms for robust community recovery require prior knowledge of the SBM parameters, creating a circular dependency between parameters and latent variables. Consequently, robustly estimating the parameters of an SBM is an important and complex problem in its own right.

To the best of our knowledge, the work of~\citet{acharya} is the first to address the problem of robust estimation on random graphs, focusing on the parameter estimation of a perturbed Erdős-Rényi model. This corresponds to the case of a single community (\(K=1\)). Their algorithm employs a greedier iterative node removal scheme that often gets trapped in poor local optima when applied to graphs with multiple communities. Our work extends to the more relevant case of \(K > 1\) where the real interest in community detection lies, and thoroughly explores the subgraph space, leading to improved solutions.

Additionally, our method provides an outlier-detection mechanism that identifies nodes most responsible for disturbing the model's quality of fit, allowing for a more nuanced analysis of outliers beyond simplistic strategies based solely on node degree distribution.

Finally, there is a notable gap in recent literature regarding experimental validation and implementation of algorithms. We provide experimental validation to our method, evaluating it across a range of synthetic data experiments to assess its performance as parameters vary, as well as performing experiments on real graphs.

\paragraph{Our Contributions}
\begin{itemize}
  \itemsep0em
  \item We prove a bound for the estimation error in Theorem \ref{thm:estim_inside_good} that generalizes the bound appearing in~\citet{acharya} to the case of graphs with multiple communities.
  \item Based on this bound, we propose a cost function to be minimized for robustly estimating the connectivity parameters of an SBM.
  \item Our main contribution is to propose an algorithm, called \textsc{SubSearch}, based on Simulated Annealing (S.A.) to minimize this cost function (Algorithm \ref{alg:find_S}). The novelty of this algorithm is that it explores the space of solutions more thoroughly, managing to escape bad local optima that previous methods fall into.
  \item \textsc{SubSearch} also serves as an outlier-detection method, identifying a set of nodes that are most deviant from the model, beyond the na\"ive pruning of nodes of extreme degree.
  \item We provide a variety of experiments to support the effectiveness of our approach, with both synthetic and real graphs. This addresses a gap in the literature of robust community detection, and to our knowledge our work is among the first to provide this experimental validation.
\end{itemize}

\subsection{Related Work}

% Acharya
The pioneer work of~\citet{acharya} estimates the parameter of an Erd\H{o}s-Rényi random graph model under node corruptions. They establish a ``certification'' bound and propose an algorithm based on iterative node removal using eigenvector scores, which can be seen as a variant of the filtering algorithm used in~\citet{diakonikolas2019robust} to robustly fit a high-dimensional Gaussian. However, they do not consider graphs with multiple communities and their algorithm requires erasing a significant portion of the original graph. Recent work by~\citet{chen2024private} extend robust estimation to inhomogeneous graphs, including SBMs. However, their approach is based on Sum-of-Squares relaxations to Semidefinite Programs (SDPs), which, despite their polynomial-time complexity in theory, may be computationally expensive in practice. In~\citet{jana2024general}, an estimator similar to the one proposed here is presented, but their approach is more akin to a direct \(k\)-means with trimming rather than the exploration of subgraph space that we do here.

Much of the prior literature focused on robust community recovery, \textit{i.e.}, accurately inferring community labels. Seminal works by~\citet{cai2015robust} and~\citet{pmlr-v49-makarychev16} explored this problem using semidefinite programs (SDPs), while~\citet{stephan2019robustness} and~\citet{Abbe_2020} study robustness of algorithms based on the spectrum of the adjacency matrix (or matrices related to it). Previously~\citet{ding_reaching_2023} studied robust recovery up to its fundamental limit on the SBM with sparse connections, using the Sum-of-Squares paradigm. The work of~\citet{srivastava_robust_2021} considers the problem of robust recovery in the more general case of sub-gaussian mixtures, and their approach is based on a linear programming relaxation of a robust SDP.

\section{SETUP}\label{setup}

\paragraph{Notation.} In what follows, \(n\) and \(K\) are positive integers with \(K \leq n\). For any finite set \(C\), \(\lvert C \rvert\) denotes the number of elements in it. All vector norms are the Euclidean norm \(\lVert x \rVert = \sum_{i=1}^n x_i^2\), and matrix norms are the spectral norm, also called the operator norm, defined for \(A \in \mathbb{R}^{m \times n}\) as \(\Vert A \Vert \coloneqq \sup_{x \in \mathbb{R}^n \setminus \{0\}} \Vert A x \Vert / \Vert x \Vert\).
It can be shown that \(\Vert A \Vert = \sqrt{\lambda_{\text{max}}(A^t A)}\), where \(\lambda_{\text{max}}(A^t A)\) denotes the maximal eigenvalue of \(A^t A\). We write \(g(n) = O(f(n))\) when there exist some positive real \(C\) and some positive integer \(n_0\) such that \(\lvert g(n) \rvert \leq C \cdot f(n)\) for all \(n \geq n_0\).
% We say \(g(n) = \Omega(f(n))\) when \(f(n) = O(g(n))\).

\paragraph{Graphs.} A graph is a pair of sets \(G = (V, E)\), where \(V = \{1, \dotsc, n\}\), is called the set of nodes and \(E \subset V \times V\) is called the set of edges. We will deal with undirected simple graphs, meaning that \((i, j) \in E \Rightarrow (j, i) \in E\) and that \((i, i) \not\in E\) for all \(i\). Such a graph can be represented in matricial form by a symmetric matrix called an adjacency matrix, defined as
\begin{equation*}
    A_{i j} =
    \begin{cases}
        \, 1 \quad \text{if } (i, j) \in E \\
        \, 0 \quad \text{otherwise}.
    \end{cases}
\end{equation*}
The degree of a vertex \(i\) is the number of edges connected to it, \textit{i.e.}, \(\textup{deg}(i) = \sum_{j=1}^n A_{ij}\).

A graph with communities is simply a graph along with a partition of its set of nodes \(V\) into \(K\) non-empty sets \((\Omega_k)_{k = 1, \dotsc, K}\) called communities, \textit{i.e.}, \(V = \Omega_1 \cup \cdots \cup \Omega_K\) with \(\Omega_i \cap \Omega_j = \emptyset, \forall i \neq j\). Communities can be represented by a community assignment vector \(z \in \{1, \dots, K\}^n\) or by a community assignment matrix \(Z \in \{0, 1\}^{n \times K}\) such that \(\sum_j Z_{ij} = 1\) for every \(i \in \{1, \dots, n\}\).

A popular generative model for graphs with communities is called the Stochastic Block Model (SBM)~\citep{HOLLAND1983109}. It has a community size parameter \(\Pi = (\pi_1, \dotsc, \pi_K)\) such that \(\forall k, 0 < \pi_k < 1\) and \(\sum_k \pi_k = 1\), and connectivity parameters \(\Gamma \in [0, 1]^{K \times K}\). Given these parameters, the SBM with \(K\) communities is a probability \(\mathbb{P}\) over the space of graphs with communities determined by
\begin{align*}
    \mathbb{P} (Z) &= \prod_{k=1}^K \pi_k^{\lvert \Omega_k \rvert}, \\
    \mathbb{P} (A \vert Z) &= \prod_{i \neq j} \Gamma_{z_i z_j}^{A_{ij}} (1 - \Gamma_{z_i z_j})^{1 - A_{ij}}.
\end{align*}
Given community assignments \(Z\), one can show that \( \mathbb{E}[A] = Q - \textup{diag}(Q)\), where \(Q = Z \Gamma Z^t\) and \(\textup{diag}(Q)\) is the \(n \times n\) containing only the diagonal of \(Q\) on the diagonal and zeroes elsewhere.

\paragraph{Submatrices.} Let \(A\) be any \(n \times n\) matrix, \(S_1, S_2 \subset \{1, \dots, n\}\) be subsets of the row and column indices, respectively. Without loss of generality, we assume that \(S_1\) is sorted and we denote \(S_1 (i)\) the \(i\)-th element of \(S_1\) (respectively for \(S_2\)). The restriction of \(A\) to \(S_1 \times S_2\) is the matrix \(A_{S_1 \times S_2}\) given, for \(i = 1, \dots, \lvert S_1 \rvert\) and \(j = 1, \dots, \lvert S_2 \rvert\), by \((A_{S_1 \times S_2})_{ij} = A_{S_1(i) S_2(j)}\). When \(S_1 = S_2 = S\), we will simply note \(A_{S \times S}\) as \(A_S\).

If \(A\) is an adjacency matrix and \(S_1, \dots, S_K\) are disjoint subsets of \(\{1, \dots, n\}\), we can estimate the connectivity parameters associated to them, for \(k, l \in \{1, \dots, K\}\), by
\begin{equation*}
    \hat{\Gamma}_{k l} = \frac{1}{\vert S_k \vert \vert S_l \vert} \sum_{i=1}^{\vert S_k \vert} \sum_{j=1}^{\vert S_l \vert} (A_{S_k \times S_l})_{i j},
\end{equation*}
defining a \(K \times K\) matrix \(\hat{\Gamma}\). This can be extended to an \(n \times n\) matrix \(\hat{Q}(S) \coloneqq \mathbf{S} \hat{\Gamma} \mathbf{S}^t\), where \(\mathbf{S}\) is the \(\vert S \vert \times K\) matrix such that \(\mathbf{S}_{ij} = 1\) if \(S(i) \in S_j\) and \(0\) otherwise.
% Notice that for \(k = l\), \(\hat{\Gamma}_{k k}\) is slightly different from the empirical edge density.

\paragraph{Clustering.}
Given an adjacency matrix \(A\), the symmetric normalized Laplacian matrix associated to it is defined as~\citep{von2007tutorial} the matrix \(L \in \mathbb{R}^{n \times n}\) with entries
\begin{equation*}
    L_{i j} = \begin{cases}
        1 \quad &\text{if } i = j \text{ and } \textup{deg}(i) \neq 0, \\
        - \frac{1}{\sqrt{\textup{deg}(i) \, \textup{deg}(j)}} \quad &\text{if } i \neq j \text{ and } A_{ij} = 1 \\
        0 \quad &\text{otherwise.}
    \end{cases}
\end{equation*}
Spectral clustering~\citep{shi2000normalized} is a widely used clustering technique that consists on applying the \(K\)-means algorithm~\citep{macqueen1967some} to the rows of the matrix whose columns are the normalized eigenvectors corresponding to the \(K\) smallest non-zero eigenvalues of \(L\). We assume that the number of clusters \(K\) is known, and for the rest of the paper all clustering is performed using spectral clustering with this given \(K\).

\paragraph{Problem Statement.}
We consider the node adversary perturbation model, where an adversary receives a sample \((Z, A_0)\) of an SBM and is allowed to arbitrarily modify the adjacency of up to \(\gamma n\) nodes, where \(\gamma \in [0, 1/2)\) is a known parameter representing the amount of corruption. This leads to the observation of a corrupted adjacency matrix \(A\). The nodes whose adjacencies were directly modified by the adversary are called outlier nodes, while the rest are called inlier nodes. We denote the set of inlier nodes as \(F\). The goal is to accurately estimate the connectivity\footnote{An adversary can bias the size parameters while being undetectable: it suffices for it to resample the connections of nodes in one community as if they were in another. For this reason, we focus on the connectivity parameters.} parameters \(\Gamma\) of the original SBM from \(A\), in the sense of minimizing the empirical estimation error \(\sum_{kl} \lvert \Gamma_{kl} - \hat{\Gamma}_{kl}\rvert\).

\section{MAIN RESULTS}
Our method is based on directly optimizing a bound relating the error obtained by estimating the connectivity parameters on a subgraph \(S\) to the spectral norm \(\lVert A_S - \hat{Q}(S) \rVert\) associated to it.

\subsection{Error Bound}
In the following result, we generalize a bound found in~\citet{acharya} to the \(K > 1\) case. The proof is in the supplementary material.

\begin{theorem}\label{thm:estim_inside_good}
Let \(A\) be an adjacency matrix sampled from a \(\gamma\)-corrupt \textup{SBM} with \(K\) communities \((\Omega_k)_{k=1, \dotsc, K}\), connectivity parameters \(\Gamma\), and inlier nodes \(F\). Furthermore, let \(S_1, \dots, S_K\) be non-empty disjoint subsets of \(\{1, \dots, n\}\), \(S\) be their union, and \(\hat{Q}(S)\) be the estimation of the expected adjacency matrix restricted to \(S\). Then,
\begin{align*}
\nonumber &\sum_{k=1}^K \sum_{l=k}^K \lvert \Gamma_{kl} - \hat{\Gamma}_{kl}\rvert \leq \frac{K^2}{\min_k{\lvert \Omega_k \cap S_k \cap F \rvert}} \\ & \times \left( \max_k{\Gamma_{kk}} + \lVert A_{F} - \mathbb{E}[A]_{F} \rVert + \lVert A_{S} - \hat{Q}(S) \rVert \right).
\end{align*}
\end{theorem}

We now use this bound to motivate our proposed objective function. On the right-hand side of the bound appearing in Theorem \ref{thm:estim_inside_good}, the term \(\max_k \Gamma_{kk}\) is bounded by 1 and becomes negligible as \(n\) increases. The term \(\lVert A_F - \mathbb{E}[A]_F \rVert\) is \(O(\sqrt{(1 - \gamma)n})\), as it represents the spectral norm of the centered adjacency matrix of the inlier subgraph, which is distributed according to an SBM~\citep{lei_rinaldo}. These two terms are independent of our choice of \(S\) and its associated clustering. Now, suppose that we had a subgraph \(S\) such that \(\lVert A_S - \hat{Q}(S) \rVert = O(\sqrt{n})\), mimicking the behavior of the inlier solution, and that \(\min_k \lvert S_k \cap \Omega_k \cap F \rvert = O(n)\). Then, Theorem \ref{thm:estim_inside_good} implies an estimation error of \(O(n^{-1/2})\). This motivates minimizing \(\lVert A_S - \hat{Q}(S) \rVert\) as a criterion for selecting a subgraph \(S\). \rev{We point out that the bound in Theorem~\ref{thm:estim_inside_good} does depend on \(\gamma\), but not in the same explicit way as in the case \(K = 1\): making the dependence on \(\gamma\) explicit is harder for the \(K > 1\) case.}
% Christine: cette borne dépend bien de \gamma, mais pas de façon explicite comme dans le K=1. (Enlever la partie "despite the proof".)

\subsection{\textsc{SubSearch}: Subgraph Search with Simulated Annealing}
\label{subsection:subgraph_search_with_simulated_annealing}
The Simulated Annealing (S.A.) algorithm, introduced by~\citet{kirkpatrick1983optimization}, is inspired by the metallurgical process of annealing, where the slow cooling of a heated solid brings it to a lower-energy state with fewer defects. Analogously, S.A. optimizes a given cost function \(c\) by exploring the state space with an initially high ``temperature'' parameter that is then slowly decreased, helping the algorithm to find global or near-global solutions.

In Algorithm \ref{alg:find_S}, we propose \textsc{SubSearch}, an algorithm using S.A. to explore the state space \(\mathcal{S}\) consisting of all subgraphs of size \((1 - \gamma)n\) of the input graph \(G\), in search for a subgraph \(S\) minimizing the cost function \(c(S) = \lVert A_S - \hat{Q}(S) \rVert\). The algorithm begins with a randomly selected connected subgraph \(S_{\text{current}} \in \mathcal{S}\) and initial temperature \(T_0\). At each step \(t\), the algorithm generates a Markov chain of length \(l_t\) at fixed temperature \(T_t\) in the following way. Define the neighborhood \(N(S)\) of any subgraph \(S \in \mathcal{S}\) as the subgraphs that can be obtained by swapping a node within \(S\) with an adjacent node outside of it. At each step \(l\) of the chain, a neighboring connected subgraph \(S_{\text{candidate}} \in N(S_{\text{current}})\) is proposed as a candidate for the next state (call to the \texttt{neighbor} function in Algorithm \ref{alg:find_S}). The cost difference between the current and the candidate state will be denoted \(\Delta \coloneqq c(S_{\text{current}}) - c(S_{\text{candidate}})\). The candidate is accepted with probability \(\min{\left(1, \exp\left(\Delta / T_t \right)\right)}\), in which case the state \(S_{\text{current}}\) is updated to it. After completing the Markov chain, the temperature is decreased following a geometric cooling schedule \(T_{t+1} = c T_t\), where \(c\) is the cooling rate parameter, typically close to 1. This process is repeated until either a maximum number of iterations \(t_{\text{max}}\) is reached or the maximal absolute variation of the cost over \(t_{\text{tol}}\) chains falls below a tolerance threshold \(\varepsilon\) (call to \texttt{stopping\_conditions} in Algorithm \ref{alg:find_S}). The algorithm returns the subgraph with minimal cost. The initial temperature is determined adaptively: starting with \(T_0 = 1\), a separate long Markov chain (\textit{e.g.}, with 100 states) is run, and \(T_0\) is multiplied by 1.5 until the rate of acceptance of neighboring states is close to one (call to \texttt{set\_initial\_temp} in Algorithm \ref{alg:find_S}). This heuristic aims at finding an initial temperature high enough to allow for an effective initial exploration phase without being so high as to slow the convergence. Notice that we implicitly cluster \(S\) when calculating \(c(S)\), due to \(\hat{Q}(S)\).

\begin{algorithm}[H]
\caption{\textsc{SubSearch}} \label{alg:find_S}
\begin{algorithmic}

% Inputs
\Require \(A\), \(K\), \(\gamma\), \(c\), \((l_t)_{t=0,\dotsc,t_{\text{max}}}\), \(t_{\text{max}}\), \(t_{\text{tol}}\), \(\varepsilon\).
    % \State Adjacency matrix: \(A\)
    % \State Number of communities: \(K\)
    % \State Corruption parameter: \(\gamma\)
    % \State Cooling rate: \(c\)
    % \State Markov chain lengths: \((l_t)_{t=0,\dotsc,t_{\text{max}}}\)
    % \State Stopping parameters: \(t_{\text{max}}\), \(t_{\text{tol}}\), \(\varepsilon\)
    \\
% Initialization
\State \(S_{\text{current}} \gets\) connected subgraph with \(\lvert S \rvert = (1-\gamma)n\)
\State \(S_{\text{best}} \gets S_{\text{current}}\)
\State \(T_0 \gets \texttt{set\_initial\_temp}(S_{\text{current}})\)

\For{\(t = 1, \dotsc, t_{\text{max}}\)}
    % Markov Chains
    \For{\(l = 1, \dotsc, l_t\)}
        % Propose candidate neighbor solution
        \State \(S_{\text{candidate}} \gets \texttt{neighbor}(S_{\text{current}})\)
        \State \(\Delta \gets c(S_{\text{current}}) - c(S_{\text{candidate}})\)
        % Acept it or not
        \State \(u \sim \mathcal{U}([0, 1])\)
        \State \(\texttt{accept\_prob} \gets \min{\left(1, \exp{\left(\Delta / T_t\right)}\right)}\)
        \If{\(u < \texttt{accept\_prob}\)}
            % Update Markov Chain state
            \State \(S_{\text{current}} \gets S_{\text{candidate}}\)
            \If{\(c(S_{\text{current}}) < c(S_{\text{best}})\)}
                \State \(S_{\text{best}} \gets S_{\text{current}}\)
            \EndIf
        \EndIf{}
    \EndFor
    % Update temperature
    \State \(T_{t+1} \gets c T_t\)
    \If{\(\texttt{stopping\_conditions}(\varepsilon, t_{\text{tol}})\)}
        \State break
    \EndIf{}
\EndFor

% Final calculations
\State \Return \(S_{\text{best}}\)
\end{algorithmic}
\end{algorithm}

\rev{
  We emphasize that our approach to the \(K > 1\) case goes beyond an incremental extension of existing methods for \(K=1\). In fact, when \(K > 1\), both outliers and misclustered inliers can increase \(\lVert A_S - \hat{Q}(S) \rVert\). The filtering approach proposed in~\citet{acharya}, which works for \(K = 1\), does not distinguish between outliers and misclustered inliers, resulting in an excessive removal of inliers. For this reason we propose subgraph exploration as an alternative.
}

\paragraph{Convergence Properties.}
Many works have studied the convergence properties of S.A.~\citep{henderson2003theory}. Intuitively, lowering the temperature slowly enough allows the observed sequence of states to form a ``near-stationary'' Markov chain, and the associated sequence of near-stationary distributions converges to a distribution supported on the set of global optima of the function being optimized.

To formally state this, let \(P_{S S^\prime} (T_t)\) be the probability of going from state \(S\) to state \(S^\prime\) at temperature \(T_t\), \textit{i.e.}, the probability of generating \(S^\prime\) from \(S\) and then accepting it. This defines an \(\lvert \mathcal{S} \rvert \times \lvert \mathcal{S} \rvert\) transition matrix \(P(T_t)\) at each temperature value. Given an initial probability vector \(\nu_0\) over the state space, the associated state probability vector at time \(t\) is defined as \(\nu_t \coloneqq P(T_{t-1}) \dots P(T_1) P(T_0) \nu_0\). Let \(\mathcal{S}^\star\) denote the set of global minima of \(c\). The vector \(\mathbf{e}^\star \in [0, 1]^{\lvert \mathcal{S} \rvert}\) is defined as the  vector with entries \(\mathbf{e}_S^{\star} \coloneqq 1 / \lvert \mathcal{S}^\star \rvert\) if \(S \in \mathcal{S}^{\star}\), \(0\) if \(S \not \in \mathcal{S}^{\star}\). Notice \(\mathbf{e}^\star\) is supported on the set of optimal solutions. Finally, we also denote \(M \coloneqq \max_{S \in \mathcal{S}} \max_{S^\prime \in N(S)} \lvert c(S) - c(S^\prime) \rvert\) the maximal local variation of the cost function, \(d(S, S^\prime)\) the minimal number of transitions needed to go from subgraph \(S\) to subgraph \(S^\prime\), and \(r \coloneqq \min_{S \in \mathcal{S} \setminus \mathcal{S}^\star} \max_{S^\prime \in \mathcal{S}} d(S, S^\prime)\) a quantity analogous to a measure of the ``radius'' of the state space according to \(d\).

\begin{proposition}[\cite{mitra1986convergence}]\label{thm:mitra}
    Suppose that the temperature evolves according to
    \begin{equation*}
        T_t = \frac{C}{\log(t + t_0 + 1)}, \quad t = 0, 1, \dots,
    \end{equation*}
    for some \(C > 0\) and arbitrary \(1 \leq t_0 < \infty\). If \(C \geq r M\), then, for any starting initial probability vector \(\nu_0\),
    \begin{equation*}
        \lim_{t \to \infty} \lVert \nu_t - \mathbf{e}^\star \rVert = 0.
    \end{equation*}
\end{proposition}

The cooling schedule in Proposition \ref{thm:mitra}, however, is often too slow for practical applications. The inverse-logarithmic decay not only progresses at a slow rate, requiring an exponential number of iterations to reach a desired temperature \(T\), but \(C\) might also grow rapidly with \(n\). \rev{In practice, faster cooling schedules are often successfully used despite their lack of theoretical guarantees, and in Algorithm~\ref{alg:find_S} we use a geometric cooling rate.}

\rev{
  We also note that since \(\lvert S \rvert = (1 - \gamma) n\), only the subgraph of inliers is entirely free of outliers. However, many solutions include \textit{a few} outliers, and for these subgraphs the method can still yield good parameter estimates. Intuitively, this is because outliers must ``conspire together'' to meaningfully affect the estimation error; when their number is small, they lose the ability to introduce significant bias to the estimation. See~\citet{moitra_essay} for a discussion of this intuition in the Gaussian case.
}

%Interestingly, this does not make our approach impractical. The reason is that our cost function can exhibit numerous high-quality local optima. To see this, let \(\delta \in (0, 1)\) be a slack parameter close to one, and consider subgraphs of size \(\lvert S \rvert = \delta (1 - \gamma) n\), slightly smaller than the number of inliers. The number of subgraphs \(S\) that contain no outlier is given by \(\binom{(1-\gamma)n}{\delta (1-\gamma)n}\), which is lower bounded by \((1/\delta)^{(1-\gamma) \delta n}\) and grows exponentially with \(n\). Given that \(\delta \approx 1\), we expect these solutions to be of high quality, representing well all communities while managing to avoid outliers. This observation justifies our method working in practice, even with the use of a geometric cooling schedule \(T_{m+1} = c T_m\), which offers significantly faster convergence.

\section{EXPERIMENTS}

We conducted several experiments to demonstrate the applicability of Algorithm~\ref{alg:find_S} and to compare it with other approaches. First, we apply our method to a single graph and analyze the relationship between the cost function, the estimation error, and the number of outliers in the subgraph. Next, we perform two ``multi-run'' experiments to understand how variations in specific parameters impact the algorithm. The first multi-run experiment explores the dependence of estimation error on the amount of perturbation \(\gamma\), while the second demonstrates that the cost-to-overlap ratio \(\lVert A_S - \hat{Q}(S) \rVert / \min_k \lvert S_k \cap \Omega_k \cap F \rvert\) appearing in the bound of Theorem \ref{thm:estim_inside_good} decreases with \(O(n^{-1/2})\). Finally, we test our method on a real-world graph. Additional experiments in the supplementary material include results on the political blogs network (with 1222 nodes), a comparison with the Degree-Corrected SBM, an analysis of exploration's importance, and an examination of the algorithm's intrinsic variability.

The code used is available at \url{https://github.com/leobianco/robust_estim_sbm/}. All experiments were executed on a 2.7 GHz Dual-Core Intel i5 processor and 8 GB of DDR3 RAM memory. All S.A. experiments were run with \texttt{seed}=12345, \texttt{cooling\_rate}=0.99, \texttt{n\_iters\_outer}=1000 outer iterations, Markov Chains of length \(L_t = m = \gamma n\), and the stopping condition of \texttt{n\_iters\_tolerance}=25 iterations with absolute cost variation below \texttt{tolerance}=\(10^{-4}\). \rev{For a more detailed discussion on how to choose hyperparameters, see~\citet{delahaye2019simulated}.}

\paragraph{Model for Corruptions.}
Since explicitly determining the worst-case perturbation for any given graph is challenging, we propose a perturbation model for our experiments on synthetic data as follows. Sample an uncorrupted graph from an SBM and \(m = \lfloor \gamma n \rfloor\) nodes to be the outliers. For each outlier \(i = 1, \dotsc, m\) and each \(k = 1, \dotsc, K\), draw a new connection probability between that node and nodes in community \(k\) using a Beta distribution, \textit{i.e.}, \(\tilde{\Gamma}_{ik} \sim \mathcal{B}(\alpha, \beta)\). Here, \(\alpha\) and \(\beta\) are chosen so that \(\mathbb{E}[\tilde{\Gamma}_{ik}] = \Gamma_{z(i) k}\) and that the variance is the greatest possible (with the constraint that \(\tilde{\Gamma}_{ik} \in [0, 1]\)). This procedure deteriorates empirical edge density estimates while avoiding being obvious to detect due to variations of degree of the corrupted nodes.

\paragraph{Baselines.}
We compare our method to three baselines. The oracle baseline estimates the parameters using the subgraph of inlier nodes with their true community labels. The pruning baseline clusters the graph, removes \texttt{num\_to\_prune / K} nodes with the highest and lowest degrees from each community (also any isolated nodes that result from this pruning), then reclusters the graph. Finally, the ``filtering'' method of~\citet{acharya} also aims to minimize \(c(S) = \lVert A_S - \hat{Q}(S) \rVert\), but in a manner different to ours. It starts with \(S_0 = G\) and at each step removes a node \(i_t\) from \(S_t\) sampled according to \(i_t \sim v_t^2\), where \(v_t\) an eigenvector associated with the top eigenvalue of \(A_S - \hat{Q}(S)\). Filtering will be allowed to remove \(n/2\) nodes, \textit{i.e.} up to half of the graph.\footnote{For clarity of exposition, we assume knowledge of the true \(\gamma\) for all methods to achieve the best possible estimation error. However, experiments with varying subgraph sizes demonstrate that our method outperforms all others even without assuming knowledge of \(\gamma\).}

\subsection{Single-run Experiments}
\begin{figure*}
  \centering
  \begin{subfigure}{\textwidth}
      \centering
      \includegraphics[width=\linewidth]{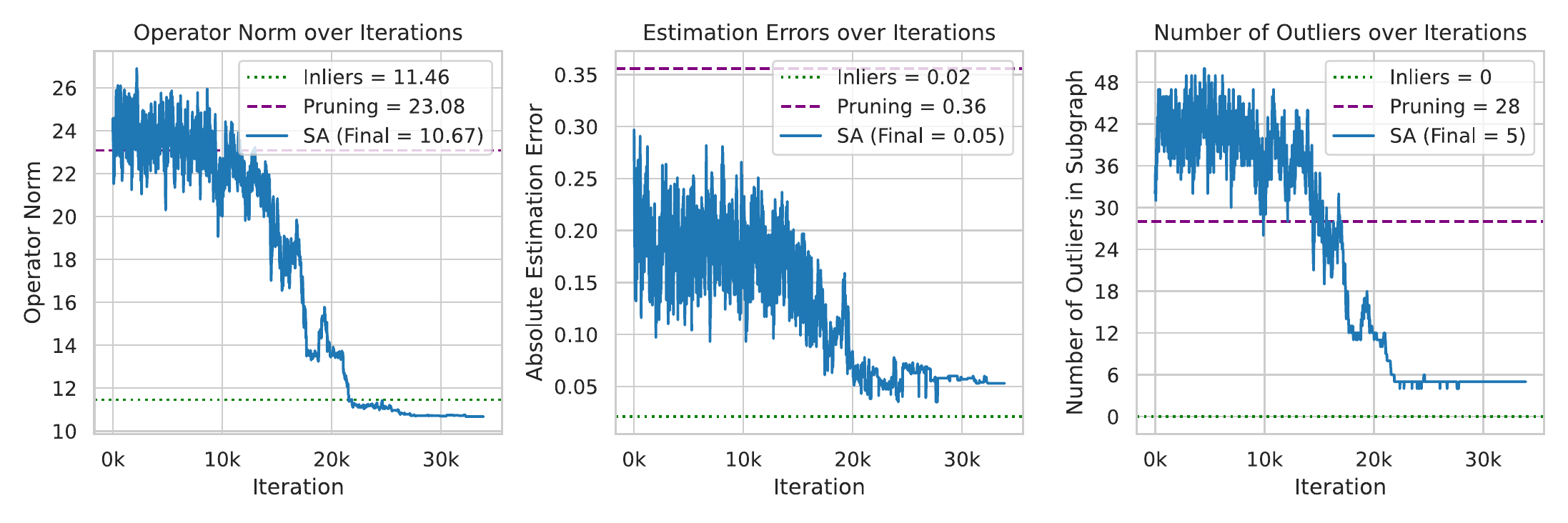}
      \caption{Results of a single-run of our method (S.A.). It decreases the cost (operator norm) and the number of outliers by exploring subgraph space and finding good solutions, while keeping subgraph size constant.}
      \label{fig:simulated_annealing_single_run}
  \end{subfigure}
  \vskip\baselineskip
  \begin{subfigure}{\textwidth}
      \centering
      \includegraphics[width=\linewidth]{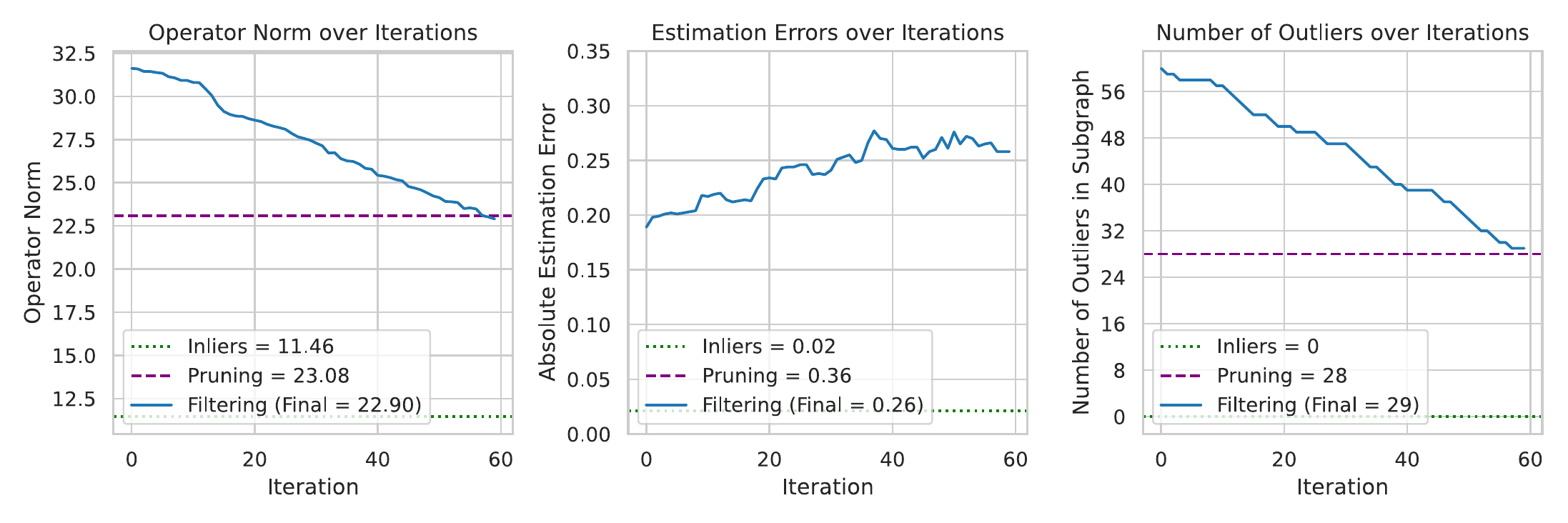}
      \caption{Results of a single-run of the competing filtering baseline. It decreases the cost (operator norm) and the number of outliers by considering smaller subgraphs at each step, but fails to decrease the error due to the lack of exploration.}
      \label{fig:filtering}
  \end{subfigure}
  \caption{Results for single-run experiments.}
  \label{fig:singlerun}
\end{figure*}

Let us analyze the behavior of a single run of our method on a perturbed graph. We consider a graph with \(n=200\) nodes, \(K=2\) communities, and connectivity parameters \(\Gamma_{11} = \Gamma_{22} = 0.65\) and \(\Gamma_{12} = 0.35\). We perturb a fraction \(\gamma=0.3\) of the nodes, \textit{i.e.}, we have \(m=60\) outliers. We will search for a subgraph with \(\lvert S \rvert = n - m = 140\) nodes with Markov chains of length \(L_t = m = 60\) for all \(t\)~\citep{van1987simulated}. Using the procedure explained in Section \ref{subsection:subgraph_search_with_simulated_annealing}, the initial temperature leading to a high initial acceptance rate was determined to be \(T_0 = 2.25\). The results are shown in Figure \ref{fig:simulated_annealing_single_run}.

% Horizontal
% \begin{figure*}
%   \centering
%   \includegraphics*[width=\linewidth]{figures/ideal_SA_horizontal.png}
%   \caption{Results of a single run of our method.}
%   \label{fig:simulated_annealing_single_run}
% \end{figure*}

% Vertical
% \begin{figure}
%   \centering
%   \caption{Single-run result of our method.}
%   \includegraphics*[width=0.85\linewidth]{figures/ideal_SA_vertical.png}
%   \label{fig:simulated_annealing_single_run}
% \end{figure}

% WITH OTHER SINGLE-RUN

The method succeeds in decreasing the spectral norm, and we see a correlation between this norm, the estimation error, and the amount of outliers inside the subgraph. The estimation error at the initial random state is \(\sum_{kl} \lvert \Gamma_{kl} - (\hat{\Gamma}_{\text{initial}})_{kl} \rvert = 0.267\), and it strongly oscillates during the first iterations as the temperature is still high and we are prone to accept moving towards a state with a worse cost. As the temperature lowers we eventually stabilize at an absolute estimation error of \(\sum_{kl} \lvert \Gamma_{kl} - (\hat{\Gamma}_{\text{SA}})_{kl} \rvert = 0.05\), and the final subgraph contains 5 outliers out of the 60.

For comparison, the error of the oracle is \(\sum_{kl} \lvert \Gamma_{kl} - (\hat{\Gamma}_{\text{inliers}})_{kl} \rvert = 0.02\). Figure \ref{fig:filtering} reveals that the filtering baseline applied on the same graph fails to decrease the estimation error, despite reducing the operator norm and number of outliers. This is because, unlike our method, filtering monotonically decreases the size of the subgraph considered, and it is this decrease in subgraph size that leads to smaller norm and less outliers. In contrast, our method keeps the subgraph size fixed and relatively large, thus the norm and outlier-count decrease due to our method finding better subgraphs rather than smaller ones. Additionally, the way filtering samples nodes to erase according to the top eigenvector is a greedier optimization technique that can lead to suboptimal solutions, whereas our S.A. exploration allows the finding of good optima. Finally, the pruning baseline with \texttt{num\_to\_prune} = \(\gamma n\) commits an error of \(\sum_{kl} \lvert \Gamma_{kl} - (\hat{\Gamma}_{\text{pruning}})_{kl} \rvert = 0.353\), much greater than the error of \(0.05\) committed by our method, and keeps 27 out of the 60 outliers, with a final cost of \(c(S_{\text{pruning}}) = 22.15\).

% Horizontal
% \begin{figure}
%   \centering
%   \includegraphics*[width=\linewidth]{figures/filtering_horizontal.png}
%   \caption{Performance of filtering.}
%   \label{fig:filtering}
% \end{figure}

% Vertical
% \begin{figure}
%   \centering
%   \caption{Single-run result of filtering baseline.}
%   \includegraphics*[height=0.5\textheight]{figures/filtering_vertical.png}
%   \label{fig:filtering}
% \end{figure}

\subsection{Multi-run Experiments}

\begin{figure}
  \centering
  \begin{subfigure}{\linewidth}
      \centering
      \includegraphics[width=\linewidth]{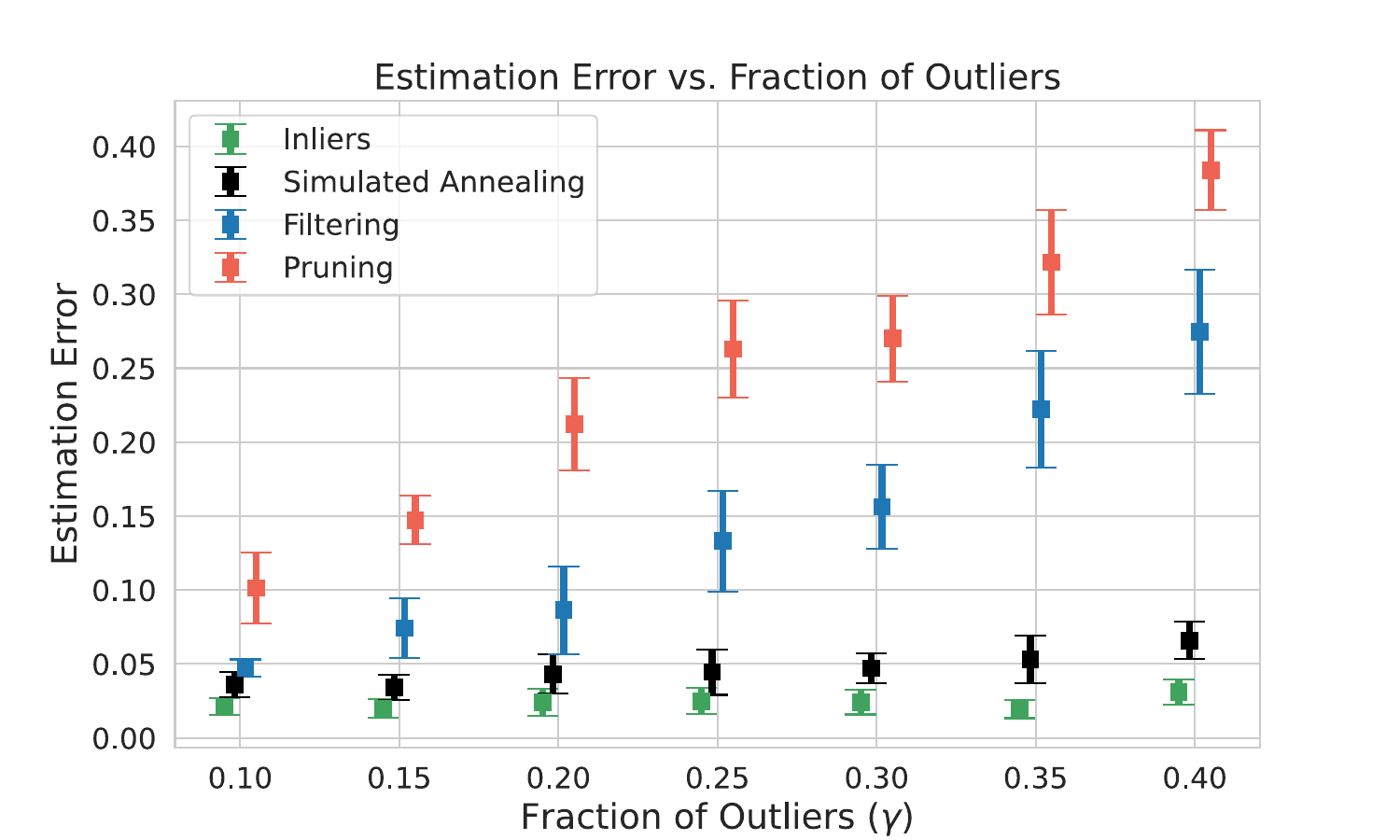}
      \caption{Estimation error of different methods as the amount of perturbation increases. Our method stays close to the inlier baseline (oracle).}
      \label{fig:error_vs_gamma_best_results}
  \end{subfigure}
  \vspace{1em}
  \begin{subfigure}{\linewidth}
      \centering
      \includegraphics[width=\linewidth]{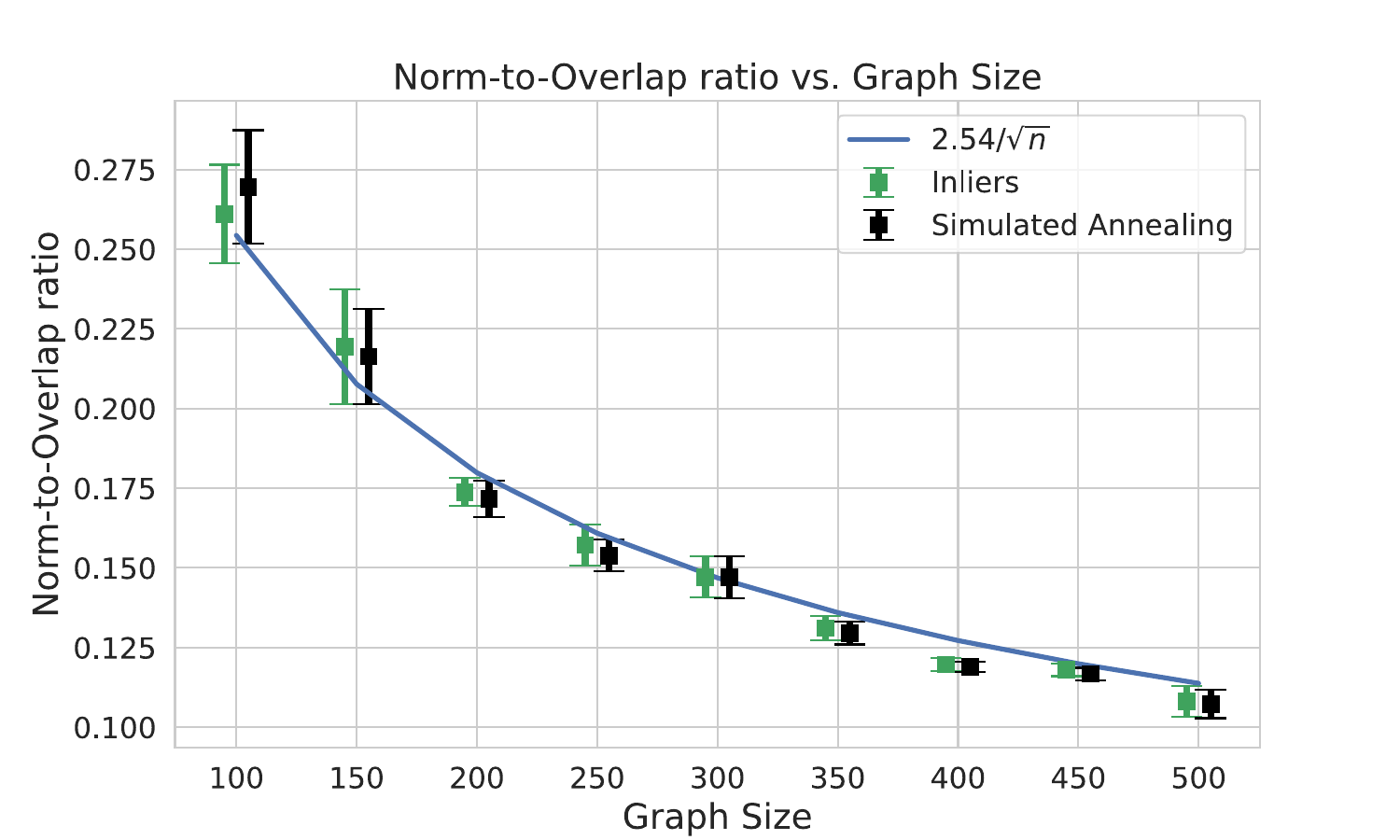}
      \caption{Cost-to-overlap ratio (defined as \(\lVert A_S - \hat{Q}(S) \rVert / \min_k \lvert S_k \cap \Omega_k \cap F \rvert\)) as \(n\) increases. As discussed after Theorem \ref{thm:estim_inside_good}, it decreases with \(O(n^{-1/2})\).}
      \label{fig:dependence_on_n}
  \end{subfigure}
  \caption{Results for multi-run experiments.}
  \label{fig:multirun}
\end{figure}

\paragraph{Dependence on Amount of Perturbation.}
We perform an experiment to study the impact of \(\gamma\) on the estimation error. We fix a grid of increasing amounts of perturbation \(\gamma = [0.10, 0.15, \dotsc, 0.40]\), then generate \(\texttt{graphs\_per\_gamma} = 10\) graphs for each amount of corruption. The estimation error is impacted by the variability of the graph generated and by the intrinsic randomness of the algorithm used. To isolate the impact due to the first of these effects from the second, we run each method \(\texttt{runs\_per\_graph} = 3\) times per graph and keep only the one achieving the least norm.\footnote{We study the variability due to the randomness of the algorithm separately, in the supplementary material.} Other parameters remain the same as in the single-run experiment. The results are shown in Figure \ref{fig:error_vs_gamma_best_results}, where the mean of the estimation error for each \(\gamma\) is represented along with a Student's 95\% confidence interval. This experiment reveals the robustness of our method, as it is the one that remains closer to the error of the oracle as the amount of corruption increases.

% ISOLATED
% \begin{figure}
%   \centering
%   \includegraphics*[width=\linewidth]{figures/multirun_gap.png}
%   \caption{Multi-run experiment showing the performance of different methods as the amount of perturbation varies.}
%   \label{fig:error_vs_gamma_best_results}
% \end{figure}

% WITH OTHER MULTIRUN - HORIZONTAL
%\begin{figure*}
%  \centering
%  \begin{subfigure}{0.48\textwidth}
%      \centering
%      \includegraphics[width=\linewidth]{figures/multirun_gamma.pdf}
%      \caption{Estimation error of different methods as the amount of perturbation increases. Our method stays close to the inlier baseline (oracle).}
%      \label{fig:error_vs_gamma_best_results}
%  \end{subfigure}
%  \hfill
%  \begin{subfigure}{0.48\textwidth}
%      \centering
%      \includegraphics[width=\linewidth]{figures/dependence_on_n.pdf}
%      \caption{Cost-to-overlap ratio (defined as \(\lVert A_S - \hat{Q}(S) \rVert / \min_k \lvert S_k \cap \Omega_k \cap F \rvert\)) as \(n\) increases. As discussed after Theorem \ref{thm:estim_inside_good}, it decreases with \(O(n^{-1/2})\).}
%      \label{fig:dependence_on_n}
%  \end{subfigure}
%  \caption{Results for multi-run experiments.}
%  \label{fig:multirun}
%\end{figure*}

% WITH OTHER MULTIRUN - VERTICAL

% \subsection{Dependence on the Connectivity Gap}
% We took \(p, q = 0.5 \pm \varepsilon\), with \(\varepsilon = [0.01, 0.04, 0.07, 0.1, 0.13, 0.16, 0.19]\). The results are in Figure \ref{fig:error_vs_gap}. When the gap is small, the quality of clustering must be bad but the quality of estimation is finally good as the parameters are close. As the gap increases, the problem shows its difficulty and we see the superiority of our method. Finally, when the gap becomes very large, the communities become easy to separate again and the error of the other baselines start decreasing again. Our method is the most stable throughout. \todo{Write.}

% \begin{figure}
%   \centering
%   \caption{Multi-run experiment showing the performance of different methods as the connectivity gap varies.}
%   \includegraphics*[width=\linewidth]{figures/multirun_gap_fast_seed_12345.png}
%   \label{fig:error_vs_gap}
% \end{figure}

% \subsection{Dependence on the Cooling Rate}
% \todo{Write.}
% This is the truly ``sensible parameter''. I was thinking of doing three single-runs to show how it impacts the solution found: \(c = 0.95 \implies\) the algorithm becomes greedy too fast and we end up in a bad local minimum \(\approx\) to a random subgraph. Then \(c = 0.98\) gives a good result. Then for \(c = 0.999\) we see that we spend too much time randomly exploring in the beginning and do not converge on time. This subsection will be very easy to write.

\paragraph{Dependence on Graph Size.}
When discussing the terms in Theorem \ref{thm:estim_inside_good}, we argued that if the cost of the solution found is \(c(S) = O(\sqrt{n})\) and the community overlap within inliers in the denominator is \(\min_k \lvert S_k \cap \Omega_k \cap F \rvert = O(n)\), then the error would decay as \(O(n^{-1/2})\). We experimentally verified this, Figure \ref{fig:dependence_on_n} shows this behavior.

% ISOLATED
% \begin{figure}
%   \centering
%   \includegraphics*[width=\linewidth]{figures/dependence_on_n.png}
%   \caption{Multi-run experiment showing the the dependence of the cost-to-overlap ratio as \(n\) increases.}
%   \label{fig:dependence_on_n}
% \end{figure}

\subsection{Application to Real Graphs}
We consider\footnote{We chose the jazz collaboration dataset to illustrate the case \(K=3\). The supplementary material contains an additional experiment on a graph of political blogs, with over a thousand nodes. We also experiment fitting a DC-SBM.} the dataset of jazz collaborations introduced in~\citet{gleiser2003community}. This graph contains 198 nodes, corresponding to jazz musicians, and 2742 edges, representing collaborations between them during the period of 1912 to 1940. Standard spectral clustering with \(K=3\) fails to properly distinguish its communities, so we turn to robust techniques.

\begin{figure}
  \centering
  \begin{subfigure}{0.45\textwidth}
      \centering
      \includegraphics[width=\linewidth]{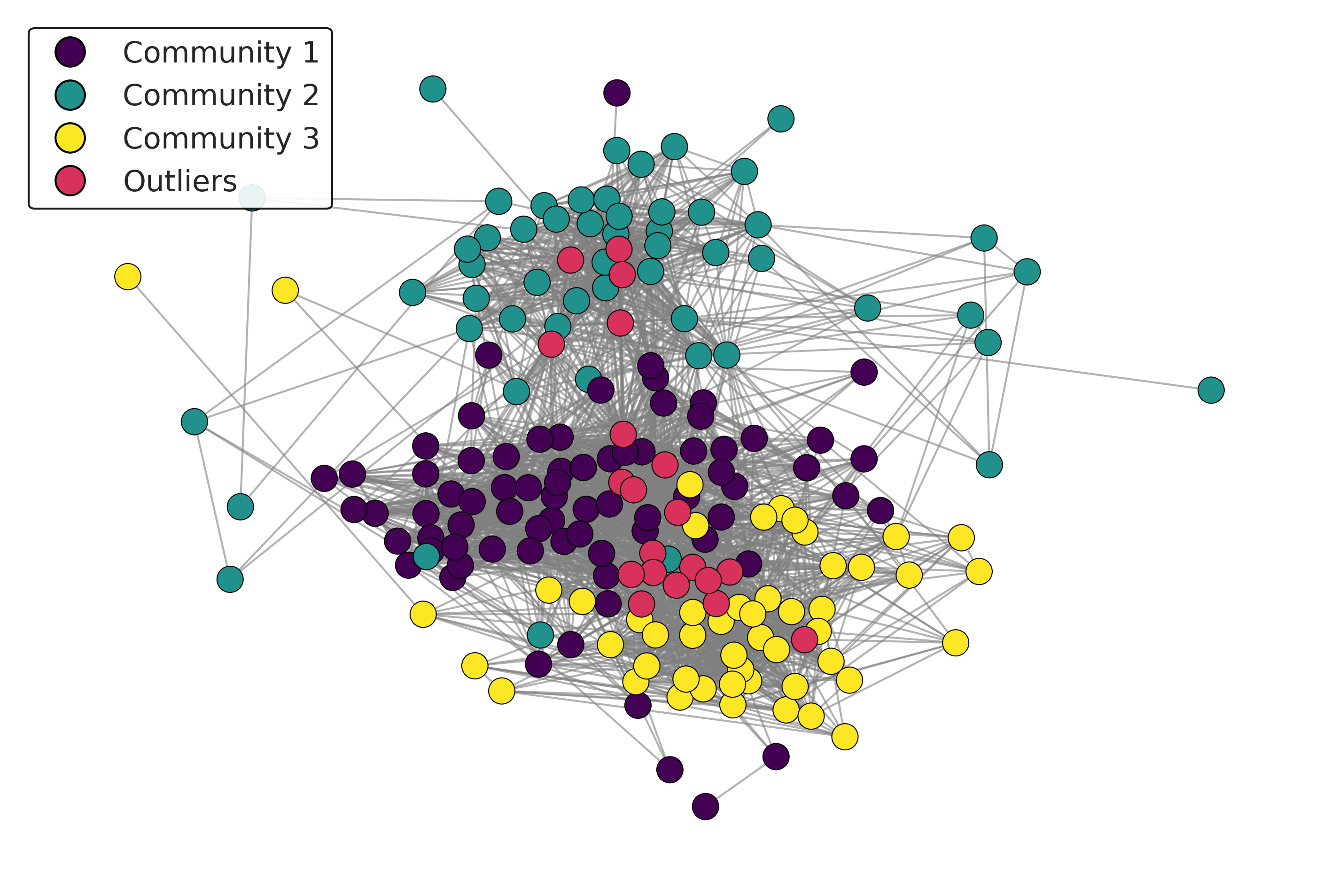}
      \caption{Community partition of the jazz collaboration graph using our method (S.A.). It identifies outlier nodes to be ignored, allowing Spectral Clustering to find the three main collaboration groups (corresponding to bands in New York, Chicago, and other cities), something it had not been capable in Figure \ref{fig:example_jazz}.}
      \label{fig:jazz_sa_graph}
  \end{subfigure}
  \vskip \baselineskip
  \begin{subfigure}{0.45\textwidth}
      \centering
      \includegraphics[width=\linewidth]{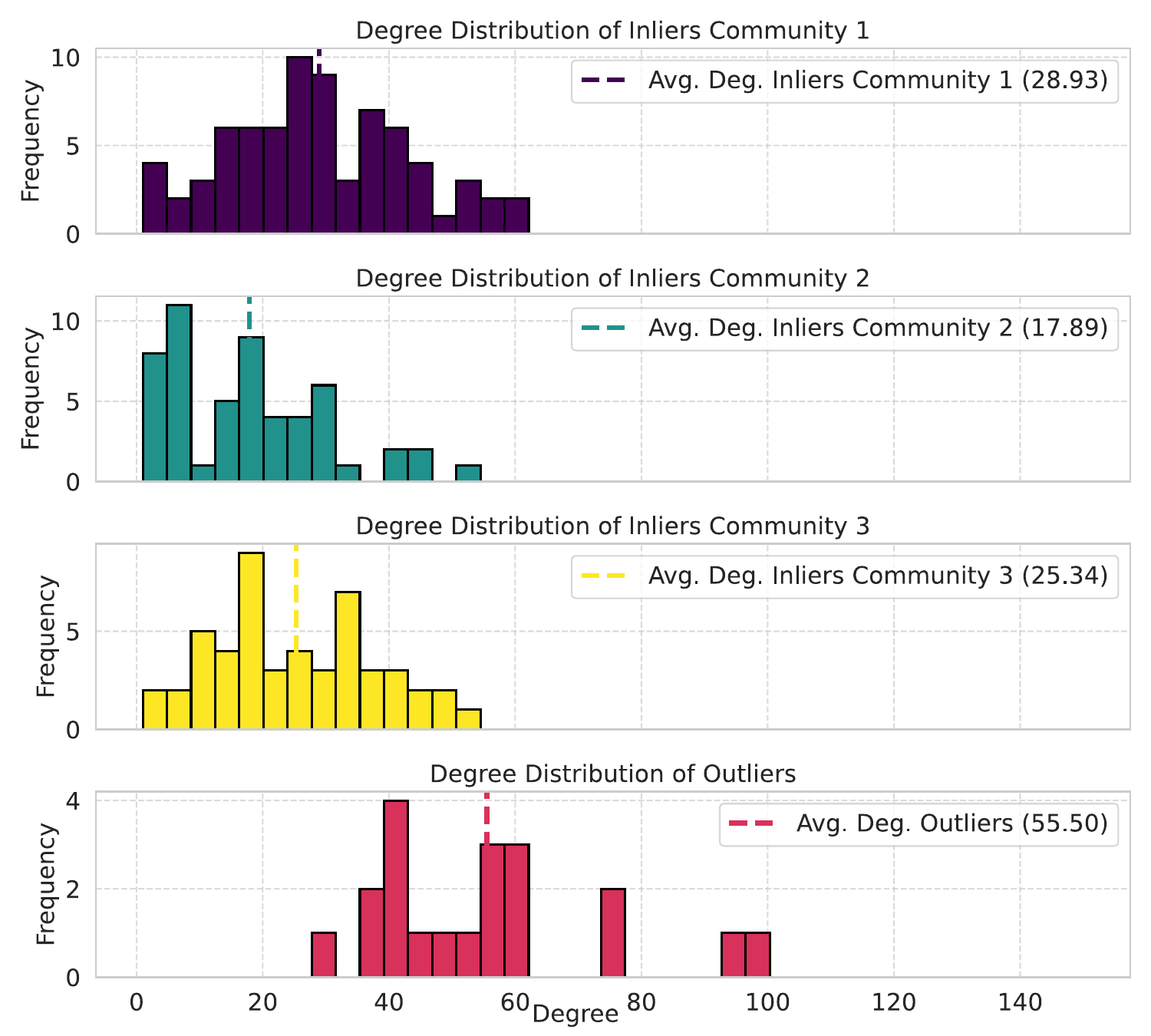}
      \caption{Degree distribution of our method (S.A.).}
      \label{fig:jazz_sa_results_2}
  \end{subfigure}
  \vskip \baselineskip
  \begin{subfigure}{0.45\textwidth}
      \centering
      \includegraphics[width=\linewidth]{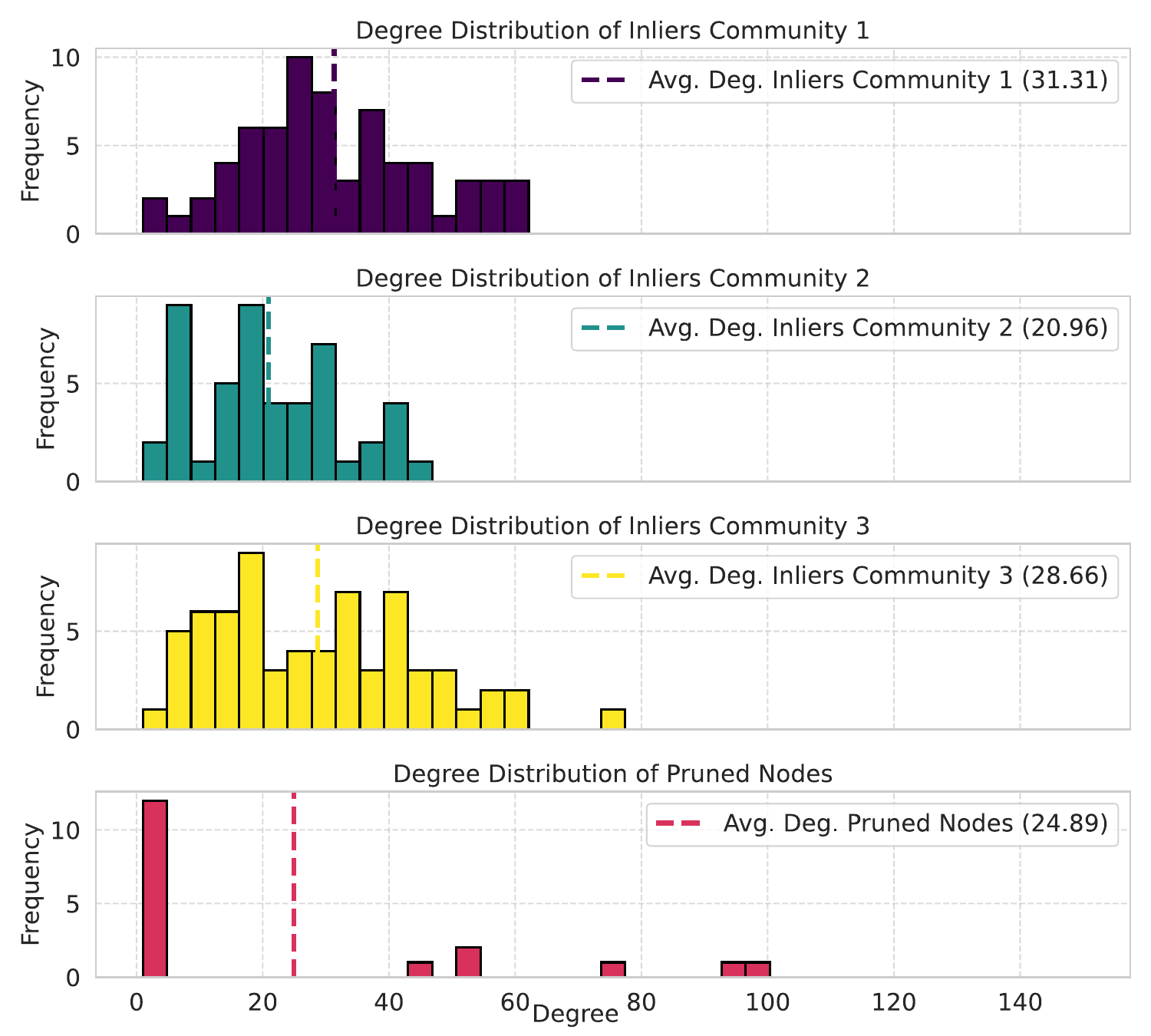}
      \caption{Degree distribution of pruning baseline.}
      \label{fig:jazz_pruning_results_2}
  \end{subfigure}
  \caption{Results for real data experiment.}
  \label{fig:multirun}
\end{figure}

We apply our method with a subgraph of size \(\lvert S \rvert = 178\), corresponding to 90\% of the size of the whole graph, and Markov chains of constant length \texttt{n\_iters\_inner}= \(\gamma n\) = 19 for each fixed temperature. The results are shown in Figures \ref{fig:jazz_sa_graph} and \ref{fig:jazz_sa_results_2}. The outlier histogram indicates that our method prunes some nodes with extreme degree, but goes beyond pruning and removes nodes of moderate-degree whose connections are not well explained by the SBM. Low-degree nodes are not removed, which might be explained by the limited node-budget the algorithm has and the impact of other outliers to the quality-of-fit being greater. The obtained estimates are \(\Gamma_{11} = 0.328\), \(\hat{\Gamma}_{12} = 0.008, \hat{\Gamma}_{13} = 0.068\), \(\hat{\Gamma}_{22} = 0.337\), \(\hat{\Gamma}_{23} = 0.017\), and \(\hat{\Gamma}_{33} = 0.351\), and the final cost is \(c(S_{\text{S.A.}}) = 14.94\).

% array([[0.328, 0.008, 0.068],
%        [0.008, 0.337, 0.017],
%        [0.068, 0.017, 0.351]])

For comparison, the pruning baseline with \texttt{num\_to\_prune}=30 yields the results in Figure \ref{fig:jazz_pruning_results_2}. Though the resulting degree distributions seem similar to those obtained before, their interpretation is different. Pruning does not detect nodes with moderate degree perturbing the estimation. The resulting estimates are \(\Gamma_{11} = 0.345\), \(\hat{\Gamma}_{12} = 0.007, \hat{\Gamma}_{13} = 0.063\), \(\hat{\Gamma}_{22} = 0.342\), \(\hat{\Gamma}_{23} = 0.017\), and \(\hat{\Gamma}_{33} = 0.332\), and the final cost \(c(S_{\text{pruning}}) = 21.11\).

\paragraph{A comment on runtimes and scalability.}
% ZACHARIE: "It is not a fair comparison, because they do not have the robustness property... they go faster but do not give a good result."
\rev{
  Due to its exploration-based nature, \textsc{SubSearch} demands a greater execution time when compared to other methods: this is a price to robustness. For instance, \textsc{SubSearch} ran the single-run experiment in 14 minutes, while filtering ran in 10 seconds, and pruning in 1 second. We highlight, however, that what we call the filtering and pruning ``baselines'' do not, in fact, exhibit the robustness properties (\textit{i.e.} low estimation error) that our method does. Thus comparing their speed with ours is not a fair comparison. Moreover, our code is not fully optimized in its current state. We believe our approach could scale to graphs with tens of thousands of nodes.
}

\section{CONCLUSION}
We introduce \textsc{SubSearch}, a method for fitting SBMs to corrupted networks while identifying potential outliers. Unlike methods that get stuck in suboptimal solutions, \textsc{SubSearch} explores the subgraph space to find high-quality fits. It identifies nodes whose connection patterns deviate from model expectations, going beyond basic outlier detection. Experiments on synthetic and real-world data show \textsc{SubSearch} is effective in fitting SBMs and spotting outliers. Future work could explore metaheuristics beyond S.A., investigate the cost function's theoretical properties for convergence and robustness, and apply the approach to other network models.

\setlength{\itemindent}{-\leftmargin}
\bibliographystyle{apalike}
\bibliography{references}

\section*{Checklist}

 \begin{enumerate}

 \item For all models and algorithms presented, check if you include:
 \begin{enumerate}
   \item A clear description of the mathematical setting, assumptions, algorithm, and/or model. [\textbf{Yes}]
   \item An analysis of the properties and complexity (time, space, sample size) of any algorithm. [\textbf{Yes}]
   \item (Optional) Anonymized source code, with specification of all dependencies, including external libraries. [\textbf{Yes}]
 \end{enumerate}

 \item For any theoretical claim, check if you include:
 \begin{enumerate}
   \item Statements of the full set of assumptions of all theoretical results. [\textbf{Yes}]
   \item Complete proofs of all theoretical results. [\textbf{Yes}]
   \item Clear explanations of any assumptions. [\textbf{Yes}]     
 \end{enumerate}

 \item For all figures and tables that present empirical results, check if you include:
 \begin{enumerate}
   \item The code, data, and instructions needed to reproduce the main experimental results (either in the supplemental material or as a URL). [\textbf{Yes}]
   \item All the training details (e.g., data splits, hyperparameters, how they were chosen). [\textbf{Yes}]
         \item A clear definition of the specific measure or statistics and error bars (e.g., with respect to the random seed after running experiments multiple times). [\textbf{Yes}]
         \item A description of the computing infrastructure used. (e.g., type of GPUs, internal cluster, or cloud provider). [\textbf{Yes}]
 \end{enumerate}

 \item If you are using existing assets (e.g., code, data, models) or curating/releasing new assets, check if you include:
 \begin{enumerate}
   \item Citations of the creator If your work uses existing assets. [\textbf{Yes}]
   \item The license information of the assets, if applicable. [\textbf{Not Applicable}]
   \item New assets either in the supplemental material or as a URL, if applicable. [\textbf{Yes}]
   \item Information about consent from data providers/curators. [\textbf{Not Applicable}]
   \item Discussion of sensible content if applicable, e.g., personally identifiable information or offensive content. [\textbf{Not Applicable}]
 \end{enumerate}

 \item If you used crowdsourcing or conducted research with human subjects, check if you include:
 \begin{enumerate}
   \item The full text of instructions given to participants and screenshots. [\textbf{Not Applicable}]
   \item Descriptions of potential participant risks, with links to Institutional Review Board (IRB) approvals if applicable. [\textbf{Not Applicable}]
   \item The estimated hourly wage paid to participants and the total amount spent on participant compensation. [\textbf{Not Applicable}]
 \end{enumerate}

 \end{enumerate}

\end{document}